\documentclass[runningheads]{llncs}
\usepackage[T1]{fontenc}
\usepackage{graphicx}
\usepackage{color}
\usepackage{hyperref}

\usepackage{amssymb}
\usepackage{amsfonts}
\usepackage{amsmath}
\usepackage[capitalize,noabbrev]{cleveref}
\usepackage{amsxtra}
\usepackage{cancel}
\usepackage{dsfont}
\usepackage{graphicx}
\usepackage{tikz}
\usepackage{tikz-qtree}
\usetikzlibrary{shapes}
\usetikzlibrary{positioning}
\usetikzlibrary{trees}
\usepackage{mathcomp}
\usepackage{mathtools}
\usepackage{multirow}
\usepackage{verbatim}
\usepackage{polynom}
\usepackage{textcomp}
\usepackage{float}
\usepackage{placeins}
\usepackage{xcolor}
\usepackage{pdflscape}
\usepackage{csquotes}
\usepackage{afterpage}
\usepackage{makecell}
\usepackage{listings}
\usepackage{url}
\usepackage{enumitem}
\usepackage{minibox}
\usepackage{algorithm}
\usepackage{algorithmic}
\usepackage{shuffle}
\usepackage{svg}
\usepackage{pifont}
\usepackage{subcaption}
\usepackage{xspace}
\usepackage{siunitx}
\usepackage{booktabs}
\usepackage{microtype}
\usepackage{footmisc}
\usepackage[export]{adjustbox}

\DeclareMathSymbol{\mlq}{\mathord}{operators}{``}
\DeclareMathSymbol{\mrq}{\mathord}{operators}{`'}

\newcommand{\R}{\mathbb R}

\newcommand{\B}{\mathcal B}

\renewcommand{\phi}{\varphi}

\newcommand{\norm}[1]{\left\lVert#1\right\rVert}
\newcommand{\abs}[1]{\left| #1 \right|}

\newcommand{\bx}{\mathbf{x}}
\newcommand{\by}{\mathbf{y}}

\newcommand{\bm}{\mathbf{m}}

\definecolor{DarkGreen}{RGB}{34,149,34}

\newcommand{\code}[1]{\texttt{#1}}
\newcommand{\cmark}{\ding{51}}%
\newcommand{\xmark}{\ding{55}}%

\newcommand{\name}{OA-CutMix}
\newcommand{\longname}{Object-Aware CutMix}
\begin{document}
\title{\name: Correcting the Label Bias of CutMix}
\author{Tobias Christian Nauen${}^{1,2}$, Stanislav Frolov${}^{2}$, Federico Raue${}^{2}$, \\ Brian B. Moser${}^2$, Andreas Dengel${}^{1, 2}$}
\authorrunning{Nauen et al.}
\institute{${}^1$ RPTU University Kaiserslautern-Landau, Kaiserslautern, Germany \\ ${}^2$ German Research Center for Artificial Intelligence (DFKI), Kaiserslautern, Germany \\ \email{first\_second.last@dfki.de} / \email{first.last@dfki.de}}
\maketitle              %

\begin{abstract}
	CutMix has become the de facto standard mixing augmentation, yet its label assignment rests on a flawed assumption: The area of the pasted patch faithfully reflects its semantic contribution to the mixed image.
	In practice, however, patches frequently land on background regions, assigning label credit to classes whose objects are not visible.
	The mean discrepancy of the CutMix label and the semantic object area is $21.5\%$.
	In $17\%$ of samples an image contributes zero visible object pixels yet receives nonzero label weight.
	We propose \longname{} (\name{}), which corrects this bias by replacing the area-based CutMix weight with one derived from precomputed segmentation masks, assigning labels in proportion to the visible object area each image contributes to the mix.
	The image mixing procedure is left entirely unchanged.
	We evaluate \name{} against 10+ static and dynamic mixing methods across 4 architectures and 6 datasets.
	\name{} consistently achieves the highest accuracy over all tasks, outperforming even dynamic mixing methods, but at a fraction of the training-time cost.
	Improvements are largest for small objects, where the label bias from CutMix is greatest.
	Thus, correcting the label is sufficient to match or exceed the performance of methods modifying the image mixing algorithm.

	\keywords{Data Augmentation \and CutMix \and Label bias.}
\end{abstract}

\section{Introduction}

\begin{figure}[t!]
	\centering
	\resizebox{.861\textwidth}{!}{\includegraphics{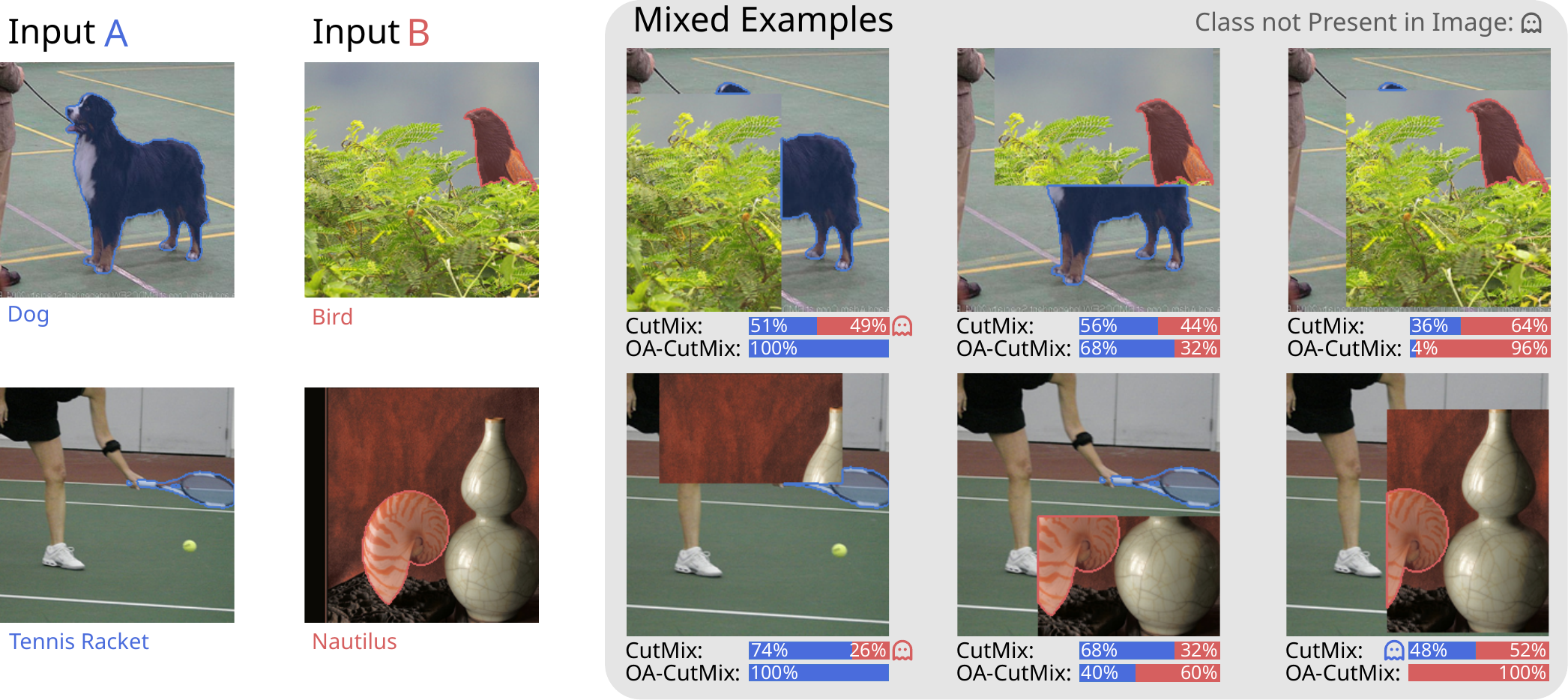}}
	\caption{CutMix labels are proportional to crop area, not semantic content, causing regression toward $0.5$, and \emph{ghost labels} (\includegraphics[width=7pt, valign=c]{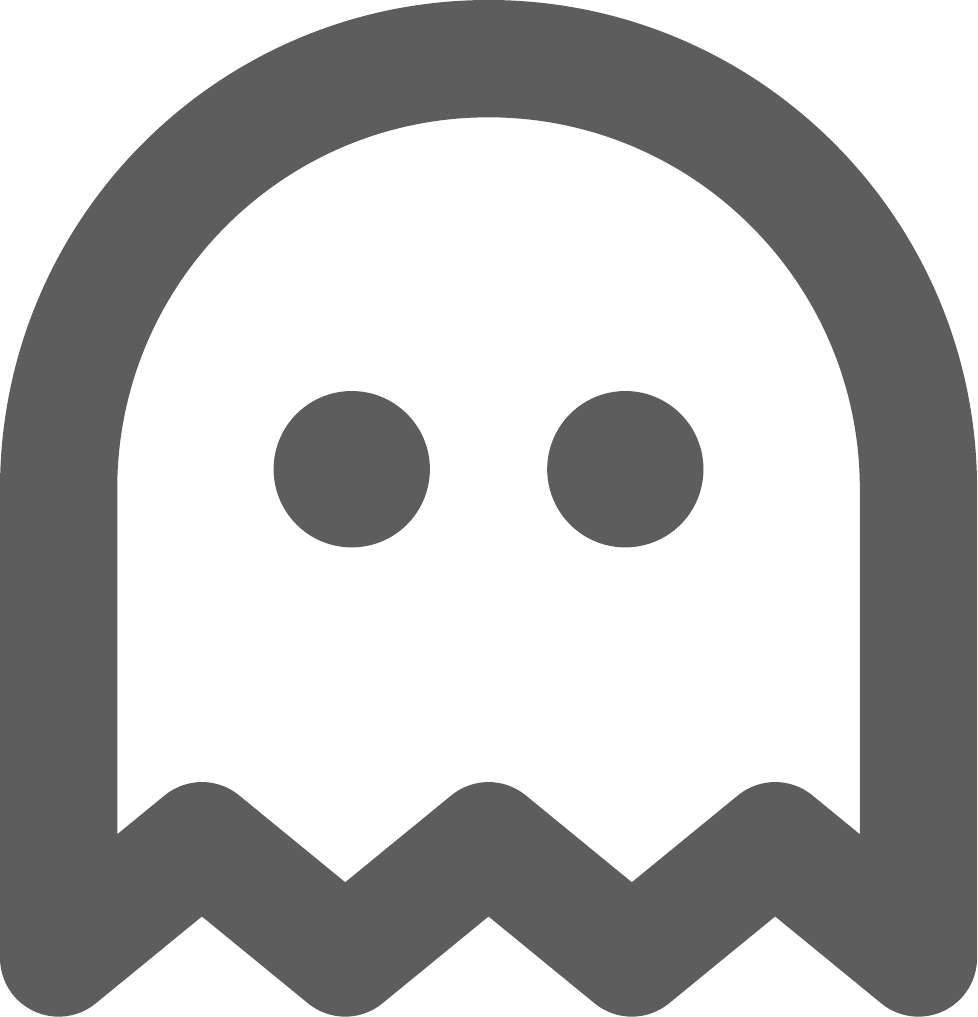}) where absent classes receive weight. Our object-aware labels track visible \emph{object} area instead.}
	\label{fig:cutmix-vs-oa-labels}
\end{figure}

Data augmentation is a fundamental technique for training vision models, improving generalization by increasing the diversity of the training data without requiring additional images~\cite{Shorten2019,Xu2023d}.
Among augmentation strategies, mix-based methods, which construct new training samples by combining pairs of images, have proven particularly effective~\cite{Yun2019,Zhang2018a}.
CutMix~\cite{Yun2019}, which pastes a rectangular patch from one image onto another and mixes labels proportionally to the patch area, has become the de facto standard, serving as a core component in state-of-the-art training recipes across a wide range of tasks~\cite{Kim2025,Liu2024b,Touvron2021b,Touvron2022}.

Despite this widespread adoption, CutMix rests on a flawed assumption: The area of the pasted patch faithfully reflects its semantic contribution to the mixed image.
In practice, however, patches frequently land on background regions, assigning label credit to a class whose object is not actually visible.
As we find in \Cref{sec:cutmix-label-problem}, this bias is both systematic and large:
The mean deviation between the CutMix label and the semantically correct object area is $21.5\%$, and in $17\%$ of CutMix samples an image contributes zero visible \emph{object} pixels yet receives nonzero label weight, a phenomenon we term \emph{ghost labels} (see \Cref{fig:cutmix-vs-oa-labels}).
For the smallest $20\%$ of objects, ghost labels even occur in $33\%$ of samples.

A parallel line of work addresses noise in CutMix through \emph{dynamic mixing}.
These methods modify the \emph{image} by selecting more salient regions to paste, so that the mixed image contains more object information and is thus more informative during training~\cite{Kim2020,Li2021,Uddin2020,Walawalkar2020}.
We hypothesize that a primary mechanism behind these improvements is not the change to the image itself, but rather the resulting improvement in label alignment.
By selecting object-rich patches, these methods incidentally reduce the label error.
To test this, we invert the approach: Rather than modifying the image to fit the label, we modify the label to fit the image.

We propose \longname{} (\name{}), an object-aware label reweighting strategy for CutMix.
Using SAM3~\cite{Carion2025}, we precompute segmentation masks for each training image.
At mixing time, we count the visible \emph{object}-pixels from each image in the mixed result and assign labels proportionally, replacing CutMix's patch-area based label.
Since masks are precomputed offline and the image mixing is identical to CutMix.
\name{}, in contrast to dynamic methods, introduces no overhead during training.

\name{} consistently outperforms 10+ static and dynamic mixing baselines across architectures and datasets.
For DeiT, \name{} outperforms all competing methods across settings.
For ResNet, it achieves the best performance over all setting when training from scratch and ranks first in all finetuning settings.
This supports our hypothesis: Correcting the label is sufficient to match or exceed the gains of methods modifying the image mixing.

\smallskip
\noindent In summary, our contributions are:
\begin{itemize}[topsep=0pt]
	\item A systematic empirical analysis of CutMix label bias, quantifying its dependence on object size and the ghost label phenomenon (\Cref{sec:cutmix-label-problem}).
	\item \name{}, a foreground-aware label reweighting strategy based on SAM3 segmentation masks, requiring no changes to the image mixing procedure and no training-time overhead (\Cref{sec:method}).
	\item A comprehensive evaluation against 10+ mixing methods across multiple architectures and datasets, supporting for the hypothesis that label correction alone recovers the gains of dynamic mixing methods (\Cref{sec:experiments}).
\end{itemize}

\section{Related Work}

\textbf{Data Augmentation.}
Data augmentation is a crucial technique for improving model performance and generalization.
Traditional augmentation strategies rely on geometric or color-space transforms to increase data diversity without changing the semantic meaning.
These simple transformations are usually bundled in more complex policies, like AutoAugment~\cite{Cubuk2019} or RandAugment~\cite{Cubuk2020}.
Recently, novel augmentations also incorporate additional knowledge from pretrained models~\cite{Nauen2025a,Trabucco2024,Rahat2025}.
We refer to \cite{Shorten2019,Xu2023d} for a general overview of data augmentation.

\textbf{Static Mixing.}
Static mixing methods mix the information of multiple images, without considering their content.
Thus, they are fast and create highly diverse input data, but are likely to output mislabeled samples (see \Cref{sec:cutmix-label-problem}).
The first of these methods was MixUp~\cite{Zhang2018a} which linearly interpolates two images.
CutMix~\cite{Yun2019} cuts out a rectangular region and pastes it onto another image, with the label as the relative area of both image regions.
FMix~\cite{Harris2020} creates dynamic shapes based on Fourier frequencies, and ResizeMix~\cite{Qin2020} takes a whole image, sizes it down, and pastes it onto another image.
AlignMix~\cite{Venkataramanan2021} uses the Sinkhorn algorithm to mix aligned regions in feature space.
Recently, Decoupled CutMix~\cite{Liu2023e} identified that the combined label (\Cref{eq:cutmix-label}) introduces competition between mix classes in the cross-entropy loss and added a regularizer for stronger gradients.

\textbf{Dynamic Mixing.}
Most dynamic mixing policies utilize additional information during training to adjust the mixed images, to feature more salient information from the mixed images.
SaliencyMix~\cite{Uddin2020} and PuzzleMix~\cite{Kim2020} extract saliency information from the trained model to adjust the cutout region to feature more salient information, while Attentive~CutMix~\cite{Walawalkar2020} uses a pretrained model.
SAMix~\cite{Li2021} and its predecessor AutoMix~\cite{Liu2021b} learn the cutout area based on a co-trained attention mechanism on top of a moving average model.
TokenMix~\cite{Liu2022e} uses patch-based and SnapMix~\cite{Huang2021} CAM weights~\cite{Zhou2015} of the current image view to reweight the label.
All of these methods have in common that they trade additional computations in the form of forward passes to convey more high-quality information to the trained model.
Despite these advances, CutMix is still used routinely~\cite{Kim2025,Touvron2021b,Touvron2022} and often outperforms more complex policies in practice~\cite{Liu2024b}.

Our approach, \name, combines the advantages of both static and dynamic mixing methods.
By using precomputed segmentation masks, we retain the training-speed of static mixing, while being able to align the labels to the mixed images, like dynamic mixing methods.
These masks, compared to saliency, CAM, or attention weights, have the advantage of being traceable through single-image data augmentation policies, since they do not depend on the specific view of an image.
They are also object-specific, covering the object without leaking into the background, preventing the creation of \emph{ghost labels} (see \Cref{sec:cutmix-label-problem}), and do not suffer from typical biases, like the exploitation of background-object correlation.

\section{Rethinking CutMix Labels}
\label{sec:cutmix-label-problem}

\begin{figure}[t!]
	\centering
	\resizebox{.98\textwidth}{!}{\includegraphics{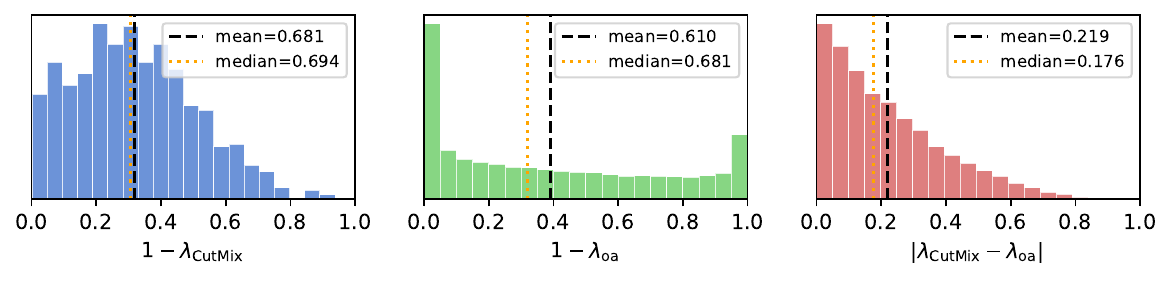}}
	\caption{Distribution of CutMix labels (left), our object-aligned labels (center), and distribution of their difference (right). Samples are often dominated by one class (see the object-aligned labels; center), which is not reflected in $\lambda_\text{CutMix}$.}
	\label{fig:lambda_dist}
\end{figure}

CutMix~\cite{Yun2019} is a data augmentation method that constructs a mixed training sample by cutting a rectangular patch from one image and pasting it onto another.
Formally, given two images $\bx_A$ and $\bx_B$ with one-hot labels $\by_A$ and $\by_B$, a box $\B = (x_1, x_2, y_1, y_2)$ is sampled within the image boundaries.
The mixed image is defined as $\tilde{x} = (1 - \mathbf{1}_{\B}) \odot \bx_A + \mathbf{1}_{\B} \odot \bx_B$ with mixed label
\begin{align}
	\tilde{y} = \lambda_{\text{CutMix}} \cdot \by_A + (1 - \lambda_{\text{CutMix}}) \cdot \by_B,
	\label{eq:cutmix-label}
\end{align}
where $\lambda_{\text{CutMix}} = 1 - \frac{(x_2 - x_1)(y_2 - y_1)}{HW}$ is the fraction of pixels retained from $\bx_A$. This patch-area based ratio is the label assignment strategy across CutMix variants.

\begin{figure}[t!]
	\begin{minipage}{.48\textwidth}
		\centering
		\resizebox{.9\textwidth}{!}{\includegraphics{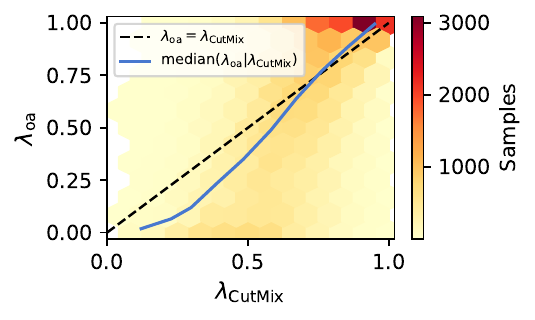}}
		\caption{2D-histogram (color) of $\lambda_\text{CutMix}$ and $\lambda_\text{oa}$. $\lambda_\text{CutMix}$ over-estimates objects inside small and outside large crops.}
		\label{fig:label-scatter}
	\end{minipage}
	\hfill
	\begin{minipage}{.48\textwidth}
		\centering
		\resizebox{.9\textwidth}{!}{\includegraphics{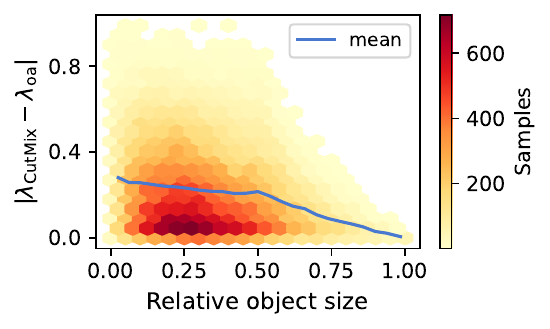}}
		\caption{Distribution of label misalignment given object size. Misalignment is larger for images with small objects.}
		\label{fig:error-vs-objsize}
	\end{minipage}
\end{figure}

The core premise of $\lambda_{\text{CutMix}}$ is that the fraction of image area occupied by a patch is a faithful proxy for its semantic contribution to the mixed image.
This assumption breaks down whenever the pasted patch contains background pixels rather than object pixels, which is a common case rather than the exception.
Using SAM3~\cite{Carion2025} to extract foreground object masks, we define $\lambda_{\text{oa}}$ as the fraction of total visible object pixels belonging to $\bx_A$ in the mixed image (\Cref{eq:oa-weight}).
\Cref{fig:lambda_dist} shows the distributions of $\lambda_{\text{CutMix}}$, $\lambda_{\text{oa}}$, and their absolute difference $\abs{\lambda_{\text{CutMix}} - \lambda_{\text{oa}}}$.
While $\lambda_{\text{CutMix}}$ clusters around its expected value, $\lambda_{\text{oa}}$ follows a different distribution, with substantially more mass near $1.0$ and $0.0$, reflecting the fact that the pasted patch frequently contributes little to no foreground content or, conversely, may cover the target image's object.
The mean absolute difference between the two is $21.5\%$, with a tail extending beyond $0.5$.
Additionally, $\lambda_{\text{CutMix}}$ and $\lambda_{\text{oa}}$ correlate only moderately ($r = 0.587$), and the median $\lambda_{\text{oa}}$ lies consistently below the diagonal for $\lambda_\text{CutMix} \leq 0.7$ (\Cref{fig:label-scatter}), indicating a systematic bias where CutMix overestimates the label contribution of the pasted image.

The bias becomes most pronounced in cases where the pasted patch contains no foreground pixels at all, yet CutMix still assigns nonzero label weight to the corresponding class.
We term these \textit{ghost labels} (see \Cref{fig:cutmix-vs-oa-labels}), as the model is asked to predict a class for which there is no visual evidence in the image.
For ImageNet, this occurs in $\mathbf{13.3\%}$ of all CutMix samples for image $B$ (the pasted image), with a mean label weight of $15.3\%$.
The symmetric case, where image $A$'s object is entirely covered by $B$'s patch, occurs in $3.7\%$ of samples, with a mean weight of $45.4\%$ still assigned to the now-absent class.
For the smallest 20\% of objects, \emph{ghost labels} even occur $33\%$ of the time, while for the smallest $10\%$ it's already $40\%$.
Together, these cases represent a substantial fraction of training samples where the soft label is not merely imprecise but categorically incorrect.

\Cref{fig:error-vs-objsize} shows the label misalignment $|\lambda_{\text{CutMix}} - \lambda_{\text{oa}}|$ as a function of object size across both images.
Images with small objects exhibit a mean label misalignment of $\approx 28\%$, while large objects that fill most of the frame approach zero misalignment, since when an object is small relative to the image, a randomly placed patch is likely to land mostly on background, making the area-based label a poor approximation of the semantic content actually present.
Conversely, when an object dominates the image, area and foreground coverage are approximately equivalent and the two labels converge.
This suggests that the label bias introduced by CutMix is not uniformly distributed across classes or datasets, but is concentrated precisely in the settings where accurate label assignment matters most: small, fine-grained objects where inter-class distinctions are subtle.

The foregoing analysis points to a straightforward remedy:
Replace $\lambda_{\text{CutMix}}$ with $\lambda_{\text{oa}}$ at label assignment time, while leaving the image mixing procedure unchanged.
This requires no change to the mixed image and no additional hyperparameters, only a semantics-aware reweighting of the soft label.

\section{\longname}
\label{sec:method}

\longname{} (\name{}) is a novel label assignment strategy for CutMix that assigns labels based on the visible object area in the combined image, rather than the area of the patch.
Concretely, \name{} uses the same rectangular patch mixing as CutMix, but replaces the area-based label weight $\lambda_\text{CutMix}$ with an object-aware one $\lambda_\text{oa}$~(\Cref{eq:oa-weight}), like in \Cref{fig:cutmix-vs-oa-labels}.
\name{} consists of two components: offline mask generation and online label reweighting.

\textbf{Mask Generation.}
\label{sec:mask-generation}
We use SAM3~\cite{Carion2025} to segment the class object in each training image prior to training, using the class label to guide segmentation.
Concretely, we use the prompt \texttt{``a <class name>, <hypernym>''} to increase the sensitivity of segmentations.
For ImageNet-based datasets, we derive the hypernym from the WordNet hierarchy; for CIFAR, we use the coarse category name and for fine-grained datasets, we use the global object type (for example \texttt{``bird''} for CUB200).
Since SAM3 benefits from higher-resolution inputs, we upscale images whose long side is smaller than $224$ pixels before segmentation and rescale the resulting mask back to the original resolution.

To handle varying detection confidence across images, we adopt an adaptive thresholding strategy.
Let $t_\text{max}(\bx)$ be the maximum detection score produced by SAM3 for a given image.
We include all detections above the threshold
\begin{align}
	t^*(\bx) = \max(0.01,\, \min(t_\text{max}(\bx),\, 0.5)),
	\label{eq:adaptive-threshold}
\end{align}
which ensures that confidently detected objects (with $t_\text{max}(\bx) \geq 0.5$) are segmented at a strict threshold, while weaker detections are retained at a relaxed threshold down to $0.01$.
All binary masks above $t^*$ are merged via a union operation, so that multiple objects of the same class are included.
If SAM3 produces no detection even at $t^* = 0.01$, no mask is created for that image; its handling during label reweighting is described below.
In practice, for ImageNet $80.9\%$ of images already yield a confident detection at $t^* = 0.5$, while only $1.6\%$ yield no detection at $t^* = 0.01$.
The result is a binary segmentation mask for each image in the dataset, stored offline and loaded alongside images during training.
Generating masks for the full TinyImageNet training set takes $152.1$ min on a single H100 GPU ($657.4$ img/s), which is less than one standard training run with CutMix ($223.1$ min).
The preprocessing cost scales linearly in the number of samples.
Since masks are computed once and reused across all experiments, this cost is amortized over subsequent runs and becomes negligible in practice.
For convenience, we release our generated masks.\footnote{\url{https://huggingface.co/datasets/TNauen/Sam3-ImageNet-Segmentations}}

\textbf{Label Reweighting.}
\label{sec:label-reweighting}
After having precomputed segmentation masks, we use these to create appropriate labels for CutMix images.
Masks are transformed jointly with their corresponding images throughout the augmentation pipeline, preserving spatial alignment across geometric transformations.

For a pair of images $\bx_A, \bx_B \in \R^{H \times W \times 3}$ with masks $\bm_A, \bm_B \in \{0, 1\}^{H \times W}$ and labels $\by_A, \by_B \in \R^C$, we sample a random box $\B$, represented by a binary mask $M_\B \in \{0,1\}^{H \times W}$.
The mixed image follows the standard CutMix formulation:
\begin{align}
	\tilde{\bx} = (1 - M_\B) \odot \bx_A + M_\B \odot \bx_B.
	\label{eq:cutmix-image}
\end{align}
Rather than weighting labels by the area of $\B$, we instead weight them by the visible object area contributed by each image.
We propose two variants for computing the visible object scores $f_A$ and $f_B$.

\noindent
\textit{Absolute mode:}
In \name/abs, we count the number of visible object pixels visible from each image after mixing:
\begin{align}
	f_A = \norm{(1 - M_\B) \odot \bm_A}_1, \qquad
	f_B = \norm{M_\B \odot \bm_B}_1.
	\label{eq:absolute-mode}
\end{align}
For images without a detected mask, we set $\bm \equiv 0$, contributing no object signal.

\noindent
\textit{Relative mode:}
\name/rel normalizes each object's count by its total size in the respective image, placing equal emphasis on objects regardless of their size:
\begin{align}
	f_A = \frac{\norm{(1 - M_\B) \odot \bm_A}_1}{\norm{\bm_A}_1}, \qquad
	f_B = \frac{\norm{M_\B \odot \bm_B}_1}{\norm{\bm_B}_1}.
	\label{eq:relative-mode}
\end{align}
For images without a detected mask, we set $\bm \equiv 1$, distributing influence uniformly over the image.
Under this convention, if neither image has a detected mask, $f_A$ and $f_B$ reduce to $\lambda_\text{CutMix} = \frac{\norm{M_\B}_1}{H \times W}$ and $1 - \lambda_\text{CutMix}$ respectively, and \name{} falls back exactly to standard CutMix labels.

Given $f_A$ and $f_B$ (absolute or relative) the object-aware mixing weight is
\begin{align}
	\lambda_\text{oa} = \frac{f_A}{f_A + f_B}
	\label{eq:oa-weight}
\end{align}
and the object-aware label is $\tilde \by_\text{oa} = \lambda_\text{oa} \by_A + (1 - \lambda_\text{oa}) \by_B$.
By construction, \name{} assigns a weight of $0$ whenever the pasted patch contains no object pixels, eliminating the ghost labels identified in \Cref{sec:cutmix-label-problem}.

\section{Experiments}
\label{sec:experiments}

\begin{table}[t!]
	\centering
	\caption{Top-1 accuracy when training DeiT from scratch on CIFAR-100, ImageNet-200, and TinyImageNet for 400, 200, and 400 epochs respectively. Latency is the time for mixing one batch of size 128 when using DeiT-S. \name{} in relative mode consistently outperforms all other mixing methods, including dynamic ones, while matching the latency of static methods.}
	\label{tab:deit-results}
	\resizebox{.79\textwidth}{!}{\begin{tabular}{lcccccccr}
	\toprule
	\multirow{2.5}{*}{Method}                       & \multicolumn{2}{c}{CIFAR-100} & \multicolumn{2}{c}{ImageNet200} & \multicolumn{2}{c}{TinyImageNet} & \multirow{2.5}{*}{Mean} & \multirow{1.5}{*}{latency}                                                      \\
	\cmidrule(lr){2-3}
	\cmidrule(lr){4-5}
	\cmidrule(lr){6-7}
	                                                & DeiT-Ti                       & DeiT-S                          & DeiT-Ti                          & DeiT-S                  & DeiT-Ti                    & DeiT-S              &                     & [ms]   \\
	\midrule
	SnapMix \cite{Huang2021}                        & $76.17$                       & $76.16$                         & $81.09$                          & $82.86$                 & $54.39$                    & $56.09$             & $71.13$             & 39.98  \\
	SAMix \cite{Li2021}                             & $79.72$                       & $79.48$                         & $81.24$                          & $82.12$                 & $55.46$                    & $57.91$             & $72.65$             & 88.27  \\
	PuzzleMix \cite{Kim2020}                       & $78.32$                       & $77.27$                         & $81.22$                          & $82.75$                 & $56.96$                    & $60.81$             & $72.89$             & 273.28 \\
	TokenMix \cite{Liu2022e}                        & $78.42$                       & $78.82$                         & $80.03$                          & $82.66$                 & $60.10$                    & $60.49$             & $73.42$             & 39.91  \\
	Attentive~CutMix \cite{Walawalkar2020}          & $79.49$                       & $79.92$                         & $81.76$                          & $83.70$                 & $59.95$                    & $56.57$             & $73.56$             & 56.10  \\
	\midrule
	Vanilla                                         & $66.46$                       & $66.91$                         & $73.32$                          & $74.56$                 & $42.83$                    & $46.18$             & $61.71$             &        \\
	ResizeMix \cite{Qin2020}                        & $71.85$                       & $70.80$                         & $79.72$                          & $79.68$                 & $50.50$                    & $51.45$             & $67.33$             & 1.05   \\
	Mixup \cite{Zhang2018a}                         & $72.58$                       & $73.69$                         & $77.69$                          & $79.18$                 & $49.62$                    & $53.71$             & $67.74$             & 0.73   \\
	FMix \cite{Harris2020}                          & $76.30$                       & $73.76$                         & $80.56$                          & $81.35$                 & $54.41$                    & $52.83$             & $69.87$             & 6.80   \\
	Mixup \cite{Zhang2018a} + CutMix \cite{Yun2019} & $78.75$                       & $77.99$                         & $81.24$                          & $82.51$                 & $56.74$                    & $60.36$             & $72.93$             & 0.70   \\
	CutMix \cite{Yun2019}                           & $79.83$                       & $79.14$                         & $81.78$                          & $83.43$                 & $61.08$                    & $58.86$             & $74.02$             & 0.67   \\
	CutMix + decoupled \cite{Liu2023e}              & $\underline{79.93}$           & $\underline{80.32}$             & $\underline{81.90}$              & $\underline{83.84}$     & $\underline{61.65}$        & $\underline{61.53}$ & $\underline{74.86}$ & 0.67   \\
	\name/rel                                       & $\mathbf{80.58}$              & $\mathbf{80.42}$                & $\mathbf{82.56}$                 & $\mathbf{84.18}$        & $\mathbf{62.82}$           & $\mathbf{64.94}$    & $\mathbf{75.92}$    & 1.03   \\
	\bottomrule
\end{tabular}
}
\end{table}

\begin{table}[t!]
	\centering
	\caption{Top-1 accuracy when training ResNet from scratch under the same settings as \Cref{tab:deit-results}. Latency is the time for mixing a batch of size 128 when using ResNet50. \name{} is the only static mixing method to match the performance of dynamic methods, while mixing more than $10$ times faster.}
	\label{tab:resnet-results}
	\resizebox{.79\textwidth}{!}{\begin{tabular}{lcccccccr}
	\toprule
	\multirow{2.5}{*}{Method}                       & \multicolumn{2}{c}{CIFAR-100} & \multicolumn{2}{c}{ImageNet200} & \multicolumn{2}{c}{TinyImageNet} & \multirow{2.5}{*}{Mean} & \multirow{1.5}{*}{latency}                                                      \\
	\cmidrule(lr){2-3}
	\cmidrule(lr){4-5}
	\cmidrule(lr){6-7}
	                                                & ResNet18                      & ResNet50                        & ResNet18                         & ResNet50                & ResNet18                   & ResNet50            &                     & [ms]   \\
	\midrule
	PuzzleMix \cite{Kim2020}                        & $\mathbf{80.90}$              & $81.20$                         & $84.29$                          & $\underline{87.91}$     & $57.71$                    & $\underline{65.37}$ & $\underline{76.23}$ & 301.00 \\
	SAMix \cite{Li2021}                             & $\underline{80.89}$           & $\mathbf{84.20}$                & $55.23$                          & $59.96$                 & $57.06$                    & $60.10$             & $66.24$             & 16.68  \\
	SnapMix \cite{Huang2021}                        & $79.81$                       & $79.18$                         & $84.07$                          & $87.57$                 & $59.25$                    & $64.64$             & $75.75$             & 51.49  \\
	Attentive~CutMix \cite{Walawalkar2020}          & $79.34$                       & $79.13$                         & $84.07$                          & $87.54$                 & $\mathbf{60.75}$           & $\mathbf{66.14}$    & $76.16$             & 56.36  \\
	\midrule
	Vanilla                                         & $77.27$                       & $78.09$                         & $83.17$                          & $86.81$                 & $50.11$                    & $57.37$             & $72.14$             &        \\
	Mixup \cite{Zhang2018a} + CutMix \cite{Yun2019} & $78.64$                       & $79.08$                         & $83.80$                          & $87.52$                 & $53.71$                    & $59.92$             & $73.78$             & 0.67   \\
	Mixup \cite{Zhang2018a}                         & $79.09$                       & $79.76$                         & $83.36$                          & $87.74$                 & $54.08$                    & $60.24$             & $74.05$             & 0.73   \\
	Alignmixup \cite{Venkataramanan2021}            & $80.52$                       & $81.70$                         & $83.73$                          & $87.56$                 & $54.34$                    & $59.56$             & $74.57$             & 1.98   \\
	FMix \cite{Harris2020}                          & $78.92$                       & $79.11$                         & $83.82$                          & $87.47$                 & $57.75$                    & $62.14$             & $74.87$             & 6.68   \\
	ResizeMix \cite{Qin2020}                        & $80.13$                       & $79.81$                         & $\mathbf{84.41}$                 & $88.01$                 & $56.04$                    & $62.28$             & $75.11$             & 0.62   \\
	CutMix \cite{Yun2019}                           & $79.26$                       & $79.34$                         & $84.08$                          & $87.74$                 & $59.43$                    & $61.79$             & $75.27$             & 0.61   \\
	CutMix + decoupled \cite{Liu2023e}              & $79.09$                       & $79.27$                         & $84.29$                          & $87.64$                 & $\underline{59.88}$        & $62.41$             & $75.43$             & 0.60   \\
	\name/rel                                       & $79.79$                       & $\underline{81.41}$             & $\underline{84.31}$              & $\mathbf{88.19}$        & $59.49$                    & $64.48$             & $\mathbf{76.28}$    & 1.02   \\
	\bottomrule
\end{tabular}
}
\end{table}

\textbf{Experimental Setup.}
We evaluate \name{} by training ResNet18/50~\cite{He2016} and DeiT-Ti/S~\cite{Touvron2021b} from scratch on CIFAR-100, TinyImageNet, and ImageNet-200, a subset of ImageNet~\cite{Deng2009} containing \emph{all} images from the 200 classes used in TinyImageNet, but at full resolution.
We additionally finetune models initialized from publicly available checkpoints on three fine-grained benchmarks: FGVC-Aircraft~\cite{Maji2013}, Stanford Cars~\cite{Krause2013}, and CUB-200~\cite{Wah2011}.
All methods are evaluated under a standard training recipe combining the respective mixing strategy with RandAugment~\cite{Cubuk2020}-based augmentation and using identical hyperparameters across methods.
Our implementation\footnote{\url{https://github.com/tobna/oa-cutmix}} is based on the OpenMixup~\cite{Li2022d} framework.

\textbf{Mixup Comparison.}
\Cref{tab:deit-results,tab:resnet-results} compare \name{} against 10+ mixing methods when training DeiT and ResNet models from scratch.
For DeiT, \name/rel achieves the highest accuracy in all six settings, surpassing both static and dynamic methods while incurring a latency overhead comparable to standard CutMix and remaining more than an order of magnitude faster at train-time than all dynamic alternatives.
For ResNet, the picture is more varied: No single method dominates across all settings, with \name/rel and PuzzleMix followed by Attentive~CutMix achieving the best or second-best result most frequently.
Averaged over all settings, \name/rel attains the highest mean accuracy, with PuzzleMix as a close second, but at $300\times$ train-time cost.
Even when including the preprocessing \emph{and} training time, \name{} is as fast as $1.1$ training runs with Attentive~CutMix and faster than a run with PuzzleMix.
Additionally, \name also improves model calibration from $9\%$ ECE to $4.3\%$, being the only method to reach sub-$5\%$ on average (see supplementary).

\begin{table}[t!]
	\centering
	\caption{Top-1 accuracy when finetuning on three fine-grained benchmarks. Each run is initialized from the same publicly available checkpoint. \name{} achieves the highest mean accuracy, with its four variants occupying the top positions.}
	\label{tab:finetuning-results}
	\resizebox{.79\textwidth}{!}{\begin{tabular}{lcccccccr}
	\toprule
	\multirow{2.5}{*}{Method}                       & \multicolumn{2}{c}{Aircraft} & \multicolumn{2}{c}{Cars} & \multicolumn{2}{c}{CUB200} & \multirow{2.5}{*}{Mean} & \multirow{1.5}{*}{latency}                                                      \\
	\cmidrule(lr){2-3}
	\cmidrule(lr){4-5}
	\cmidrule(lr){6-7}
	                                                & DeiT-S                       & ResNet50                 & DeiT-S                     & ResNet50                & DeiT-S                     & ResNet50            &                     & [ms]   \\
	\midrule
	TokenMix \cite{Liu2022e}                        & $78.70$                      &                          & $89.83$                    &                         & $80.69$                    &                     &                     & 39.91  \\
	Attentive~CutMix \cite{Walawalkar2020}          & $78.46$                      & $71.98$                  & $89.44$                    & $76.78$                 & $82.48$                    & $51.38$             & $75.09$             & 56.10  \\
	PuzzleMix \cite{Kim2020}                       & $\mathbf{82.93}$             & $76.45$                  & $\mathbf{91.21}$           & $80.62$                 & $\mathbf{84.05}$           & $59.08$             & $79.06$             & 273.28 \\
	\midrule
	Vanilla                                         & $79.99$                      & $86.02$                  & $88.92$                    & $90.16$                 & $81.69$                    & $9.44$              & $72.70$             &        \\
	FMix \cite{Harris2020}                          & $76.93$                      & $70.87$                  & $89.69$                    & $77.04$                 & $79.89$                    & $49.90$             & $74.05$             & 6.80   \\
	CutMix \cite{Yun2019}                           & $79.60$                      & $72.07$                  & $89.14$                    & $76.58$                 & $79.69$                    & $49.62$             & $74.45$             & 0.67   \\
	ResizeMix \cite{Qin2020}                        & $78.43$                      & $74.89$                  & $90.25$                    & $80.36$                 & $81.64$                    & $55.82$             & $76.90$             & 1.05   \\
	CutMix + decoupled \cite{Liu2023e}              & $80.53$                      & $86.23$                  & $89.67$                    & $90.98$                 & $80.00$                    & $81.84$             & $84.87$             & 0.67   \\
	Mixup \cite{Zhang2018a}                         & $79.84$                      & $\underline{86.80}$      & $89.47$                    & $\underline{91.00}$     & $\underline{82.52}$        & $\mathbf{83.00}$    & $85.44$             & 0.73   \\
	Mixup \cite{Zhang2018a} + CutMix \cite{Yun2019} & $80.65$                      & $86.65$                  & $90.40$                    & $90.51$                 & $82.50$                    & $82.24$             & $85.49$             & 0.70   \\
	\name/rel                                       & $\underline{82.42}$          & $85.96$                  & $\underline{90.66}$        & $\underline{91.00}$     & $82.00$                    & $82.00$             & $\underline{85.67}$ & 1.03   \\
	\name/abs                                       & $81.82$                      & $\mathbf{87.28}$         & $90.61$                    & $\mathbf{91.42}$        & $81.45$                    & $\underline{82.71}$ & $\mathbf{85.88}$    & 1.01   \\
	\bottomrule
\end{tabular}
}
\end{table}

\Cref{tab:finetuning-results} reports results for finetuning on fine-grained datasets.
While for DeiT-S, PuzzleMix is the performing best, \name{}/rel is the close second best, with the mixing operation being more than $200$ times faster.
For ResNet50, \name/abs is the best performing method, reaching first place in two and second in one out of the three settings.
Notably, dynamic methods that are competitive when training from scratch are substantially weaker when finetuning ResNet.
We also find that MixUp outperforms CutMix in the finetuning setting, which might be the result of the pretraining of the public checkpoints including Mixup, making it a natural fit during finetuning.

\begin{figure}[t!]
	\begin{subfigure}{.5\textwidth}
		\centering
		\resizebox{.7\textwidth}{!}{\includegraphics{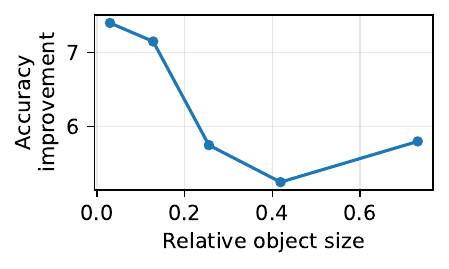}}
		\caption{DeiT-S on TinyImageNet.}
	\end{subfigure}
	\begin{subfigure}{.5\textwidth}
		\centering
		\resizebox{.7\textwidth}{!}{\includegraphics{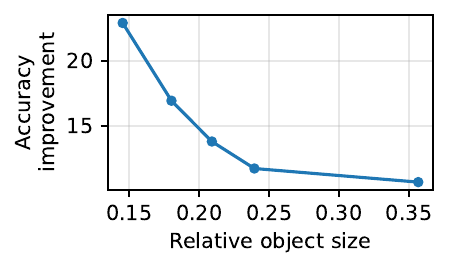}}
		\caption{ResNet50 on FGVC-Aircraft.}
	\end{subfigure}
	\caption{Accuracy improvement (p.p.) of training with \name{} over CutMix, compared to object size. The adjusted labels of \name{} are especially beneficial for images with smaller objects, precisely where the label error is largest.}
	\label{fig:acc-difference-vs-object-size}
\end{figure}

To validate the claim of \Cref{sec:cutmix-label-problem}, we partition the test sets into equal-sample bins by object size and measure the per-bin accuracy improvement of training with \name{} over CutMix.
\Cref{fig:acc-difference-vs-object-size} shows that the accuracy gain of \name{} is largest for small objects, mirroring the misalignment profile of \Cref{fig:error-vs-objsize}.
This confirms that improvements delivered by \name{} are concentrated precisely where CutMix's area-based label assignment is most misaligned.

\begin{table}[t!]
	\caption{Method ablation. \name/rel leads for training from scratch and finetuning DeiT, while \name/abs is competitive for finetuning ResNet.}
	\label{tab:oacutmix-setting-ablation}
	\centering
	\resizebox{.95\textwidth}{!}{\begin{tabular}{lcccccccc}
	\toprule
	\multirow{2.5}{*}{Method}          & \multicolumn{4}{c}{DeiT-S} & \multicolumn{4}{c}{ResNet50}                                                                                                                                     \\
	\cmidrule(lr){2-5} \cmidrule(l){6-9}
	                                   & CIFAR-100                  & ImageNet200                  & Aircraft            & CUB200              & CIFAR-100           & ImageNet200         & Aircraft            & CUB200              \\
	\midrule
	CutMix \cite{Yun2019}              & $79.14$                    & $83.43$                      & $79.60$             & $79.69$             & $79.34$             & $87.74$             & $72.07$             & $49.62$             \\
	CutMix + decoupled \cite{Liu2023e} & $80.32$                    & ${83.84}$                    & $80.53$             & $80.00$             & $79.27$             & $87.64$             & $86.23$             & $81.84$             \\
	\midrule
	\name/abs + decoupled              & $80.09$                    & $83.96$                      & $82.12$             & $81.45$             & $\underline{80.54}$ & $87.90$             & $\underline{86.86}$ & $\underline{82.67}$ \\
	\name/abs                          & $80.78$                    & $83.68$                      & $\underline{81.82}$ & $81.45$             & $80.35$             & $87.47$             & $\mathbf{87.28}$    & $\mathbf{82.71}$    \\
	\name/rel + decoupled              & $\mathbf{80.97}$           & $\underline{84.06}$          & $81.58$             & $\underline{81.71}$ & $79.05$             & $\underline{88.11}$ & $86.44$             & $82.36$             \\
	\name/rel                          & $\underline{80.42}$        & $\mathbf{84.18}$             & $\mathbf{82.42}$    & $\mathbf{82.00}$    & $\mathbf{81.41}$    & $\mathbf{88.19}$    & $85.96$             & $82.00$             \\
	\bottomrule
\end{tabular}
}
\end{table}

\textbf{Ablations.}
When comparing \name's modes, relative mode outperforms absolute mode when training from scratch (\Cref{tab:oacutmix-setting-ablation}), while absolute mode has a slight edge when finetuning ResNet (\Cref{tab:finetuning-results}).
We hypothesize that relative mode benefits from normalizing object sizes, which is more important when training from scratch on datasets with high intra-class size variation.
Combining \name{} with the decoupled loss regularizer of~\cite{Liu2023e} does not improve over standard cross-entropy loss.
This contrasts with standard CutMix labels, where the regularizer reliably helps,
suggesting that the decoupled loss primarily compensates for label inaccuracies that \name{} already corrects.

\begin{table}[t!]
	\centering
	\caption{Ablation of mask quality. Masks are progressively degenerated to isolate the contribution of positional and shape information.}
	\label{tab:mask-ablation}
	\resizebox{.95\textwidth}{!}{\begin{tabular}{lccccccccccc}
	\toprule
	\multirow{4}{*}{Method}            & \multirow{4}{*}{Masks} & \multirow{4}{*}{Pos.} & \multirow{4}{*}{Size} & \multicolumn{4}{c}{absolute mode} & \multicolumn{4}{c}{relative mode}                                                                                                                                                         \\
	\cmidrule(lr){5-8} \cmidrule(lr){9-12}
	                                   &                        &                       &                       & \multicolumn{2}{c}{CUB200}        & \multicolumn{2}{c}{TinyImageNet}  & \multicolumn{2}{c}{CUB200} & \multicolumn{2}{c}{TinyImageNet}                                                                                         \\
	\cmidrule(lr){5-6} \cmidrule(lr){7-8} \cmidrule(lr){9-10} \cmidrule(lr){11-12}
	                                   &                        &                       &                       & DeiT-S                            & ResNet50                          & DeiT-S                     & ResNet50                         & DeiT-S              & ResNet50            & DeiT-S              & ResNet50            \\
	\midrule
	CutMix \cite{Yun2019}              & \xmark                 &                       &                       & $79.69$                           & $49.62$                           & $58.86$                    & $61.79$                          & $79.69$             & $49.62$             & $58.86$             & $61.79$             \\
	CutMix + decoupled \cite{Liu2023e} & \xmark                 &                       &                       & $80.00$                           & $81.84$                           & $61.53$                    & $62.41$                          & $80.00$             & $81.84$             & $61.53$             & $62.41$             \\
	\midrule
	\multirow{5}{*}{\name{}}           & Inter-Class Mix        & \xmark                & \xmark                & $80.48$                           & $81.12$                           & $61.16$                    & $62.50$                          & $80.79$             & $81.34$             & $62.50$             & $63.39$             \\
	                                   & Intra-Class Mix        & \xmark                & $\mathbf{\sim}$       & $81.14$                           & $81.08$                           & $61.32$                    & $62.64$                          & $\underline{81.24}$ & $81.41$             & $62.05$             & $63.39$             \\
	                                   & bounding box           & \cmark                & \xmark                & $81.01$                           & $81.98$                           & $63.30$                    & $64.27$                          & $81.00$             & $\mathbf{82.10}$    & $\underline{64.04}$ & $\underline{63.96}$ \\
	                                   & SAM3                   & \cmark                & \cmark                & $\mathbf{81.45}$                  & $\mathbf{82.71}$                  & $\mathbf{63.69}$           & $\mathbf{65.53}$                 & $\mathbf{82.00}$    & $\underline{82.00}$ & $\mathbf{64.94}$    & $\mathbf{64.48}$    \\
	                                   & annotations            & \cmark                & \cmark                & $\underline{81.19}$               & $\underline{82.64}$               &                            &                                  & $80.86$             & $81.98$                                                         \\
	\bottomrule
\end{tabular}
}
\end{table}

\Cref{tab:mask-ablation} ablates the sensitivity of \name{} to mask quality by progressively degrading the masks used for label reweighting.
Replacing SAM3 masks with bounding boxes, retaining spatial position but discarding shape information, recovers roughly two thirds of the gain over the CutMix baseline, confirming that positional information is already informative.
Shuffling masks within a class further degrades performance, as masks no longer correspond to the correct image but still encode class-level shape and size statistics; shuffling across classes degrades performance further still.
Taken together, these results show a monotonic relationship between mask quality and accuracy, while also demonstrating that \name{} is robust to imperfect segmentation.
Interestingly, SAM3-generated masks outperform the human annotations available for CUB200.
We attribute this to the coarseness of these annotations~\cite{Wah2011} compared to SAM3's output quality.
Localizing the bird in an image is straightforward for SAM3 and its masks are more precise than the annotations provided in the dataset.

\section{Conclusion \& Limitations}
\label{sec:conclusion}
We have identified and quantified a systematic label bias in CutMix.
Since mixing weights are assigned by patch area rather than by visible object content, a substantial fraction of training samples carry incorrect soft labels, with the problem most acute for small objects.
To address this, we proposed \name{}, which replaces the area-based label with one derived from precomputed segmentation masks.
Experiments across four architectures and six datasets confirm that \name{} consistently improves over CutMix and matches or exceeds dynamic alternatives that are an order of magnitude slower during training.
The accuracy gain correlates directly with object size, mirroring the label error profile.
These results support our hypothesis:
The gains of dynamic mixing can be largely recovered by only aligning the label to the static-mixed image.

\textbf{Limitations.}
\name{} inherits a dependency on SAM3 for mask generation, which requires a one-time offline preprocessing step.
Additionally, performance might degrade for settings where SAM3 mask quality degrades, due to low or biased data in its training set, as reflected in our mask quality ablation.

\subsection*{Acknowledgements}
This work was funded by the Carl-Zeiss Foundation under the Sustainable Embedded AI project (P2021-02-009) and by the BMFTR project Albatross (funding code 16IW24002).
All compute was done thanks to the Pegasus cluster at DFKI Kaiserslautern.

\bibliographystyle{splncs04}
\bibliography{main}

\begin{thebibliography}{10}
\providecommand{\url}[1]{\texttt{#1}}
\providecommand{\urlprefix}{URL }
\providecommand{\doi}[1]{https://doi.org/#1}

\bibitem{Carion2025}
Carion, N., Gustafson, L., Hu, Y.T., Shoubhik, Hu, R., Suris, D., Ryali, C., Alwala, K.V., Khedr, H., Huang, A., Lei, J., Ma, T., Guo, B., Kalla, A., Marks, M., Greer, J., Wang, M., Sun, P., Rädle, R., Afouras, T., Mavroudi, E., Xu, K., Wu, T.H., Zhou, Y., Momeni, L., Hazra, R., Ding, S., Vaze, S., Porcher, F., Li, F., Li, S., Kamath, A., Cheng, H.K., Dollár, P., Ravi, N., Saenko, K., Zhang, P., Feichtenhofer, C.: Sam 3: Segment anything with concepts  (Nov 2025)

\bibitem{Cubuk2020}
Cubuk, E.D., Zoph, B., Shlens, J., Le, Q.V.: Randaugment: Practical automated data augmentation with a reduced search space. In: International Conference on Neural Information Processing Systems. Curran Associates Inc. (2020)

\bibitem{Cubuk2019}
Cubuk, E.D., Zoph, B., Man{\'e}, D., Vasudevan, V., Le, Q.V.: Autoaugment: Learning augmentation strategies from data. 2019 IEEE/CVF Conference on Computer Vision and Pattern Recognition (CVPR) pp. 113--123 (2019)

\bibitem{Deng2009}
Deng, J., Dong, W., Socher, R., Li, L.J., Li, K., Fei-Fei, L.: {ImageNet}: A large-scale hierarchical image database. In: 2009 {IEEE} Conference on Computer Vision and Pattern Recognition (2009)

\bibitem{Guo2017}
Guo, C., Pleiss, G., Sun, Y., Weinberger, K.Q.: On calibration of modern neural networks. In: International Conference on Machine Learning. PMLR (2017)

\bibitem{Harris2020}
Harris, E., Marcu, A., Painter, M., Niranjan, M., Prügel-Bennett, A., Hare, J.: Fmix: Enhancing mixed sample data augmentation  (2020)

\bibitem{He2016}
He, K., Zhang, X., Ren, S., Sun, J.: Deep residual learning for image recognition. In: IEEE Conference on Computer Vision and Pattern Recognition (2016)

\bibitem{Huang2021}
Huang, S., Wang, X., Tao, D.: Snapmix: Semantically proportional mixing for augmenting fine-grained data. AAAI Conference on Artificial Intelligence  (2021)

\bibitem{Kim2020}
Kim, J.H., Choo, W., Song, H.O.: Puzzle mix: Exploiting saliency and local statistics for optimal mixup. In: International Conference on Machine Learning (2020)

\bibitem{Kim2025}
Kim, S., Im, H., Lee, W., Lee, S., Kang, P.: Robustmixgen: Data augmentation for enhancing robustness of visual–language models in the presence of distribution shift. Neurocomputing  \textbf{619},  129167 (Feb 2025)

\bibitem{Krause2013}
Krause, J., Stark, M., Deng, J., Fei-Fei, L.: 3d object representations for fine-grained categorization. In: 4th International IEEE Workshop on 3D Representation and Recognition (3dRR-13). Sydney, Australia (2013)

\bibitem{Li2021}
Li, S., Liu, Z., Wang, Z., Wu, D., Liu, Z., Li, S.Z.: Boosting discriminative visual representation learning with scenario-agnostic mixup  (2021)

\bibitem{Li2022d}
Li, S., Wang, Z., Liu, Z., Tian, J., Wu, D., Tan, C., Jin, W., Li, S.Z.: Openmixup: Open mixup toolbox and benchmark for visual representation learning  (2022)

\bibitem{Liu2024b}
Liu, C., Fan, F., Schwarz, A., Maier, A.: Cut to the mix: Simple data augmentation outperforms elaborate ones in limited organ segmentation datasets. Medical Image Computing and Computer Assisted Intervention – MICCAI 2024 pp. 145--154 (2024)

\bibitem{Liu2022e}
Liu, J., Liu, B., Zhou, H., Li, H., Liu, Y.: Tokenmix: Rethinking image mixing for data augmentation in vision transformers. In: European conference on computer vision. pp. 455--471. Springer (2022)

\bibitem{Liu2023e}
Liu, Z., Li, S., Wang, G., Wu, L., Tan, C., Li, S.Z.: Harnessing hard mixed samples with decoupled regularizer. In: Conference on Neural Information Processing Systems -- NeurIPS (2023)

\bibitem{Liu2021b}
Liu, Z., Li, S., Wu, D., Liu, Z., Chen, Z., Wu, L., Li, S.Z.: Automix: Unveiling the power of mixup for stronger classifiers. In: European Conference on Computer Vision. pp. 441--458. Springer (2022)

\bibitem{Maji2013}
Maji, S., Kannala, J., Rahtu, E., Blaschko, M., Vedaldi, A.: Fine-grained visual classification of aircraft. Tech. rep. (2013)

\bibitem{Nauen2025a}
Nauen, T.C., Moser, B., Raue, F., Frolov, S., Dengel, A.: Foraug: Recombining foregrounds and backgrounds to improve vision transformer training with bias mitigation  (2025)

\bibitem{Qin2020}
Qin, J., Fang, J., Zhang, Q., Liu, W., Wang, X., Wang, X.: Resizemix: Mixing data with preserved object information and true labels  (2020)

\bibitem{Rahat2025}
Rahat, F., Hossain, M.S., Ahmed, M.R., Jha, S.K., Ewetz, R.: Data augmentation for image classification using generative ai. In: Proceedings of the Winter Conference on Applications of Computer Vision (WACV). pp. 4173--4182 (February 2025)

\bibitem{Shorten2019}
Shorten, C., Khoshgoftaar, T.M.: A survey on image data augmentation for deep learning. Journal of Big Data  \textbf{6}(1) (2019)

\bibitem{Touvron2021b}
Touvron, H., Cord, M., Douze, M., Massa, F., Sablayrolles, A., Jegou, H.: Training data-efficient image transformers \& distillation through attention. In: Meila, M., Zhang, T. (eds.) International Conference on Machine Learning (2021)

\bibitem{Touvron2022}
Touvron, H., Cord, M., J{\'e}gou, H.: Deit iii: Revenge of the vit. In: Computer Vision -- ECCV 2022. pp. 516--533. Springer Nature Switzerland, Cham (2022)

\bibitem{Trabucco2024}
Trabucco, B., Doherty, K., Gurinas, M.A., Salakhutdinov, R.: Effective data augmentation with diffusion models. In: International Conference on Learning Representations -- ICLR (2024)

\bibitem{Uddin2020}
Uddin, A.F.M.S., Monira, M.S., Shin, W., Chung, T., Bae, S.H.: Saliencymix: A saliency guided data augmentation strategy for better regularization. International Conference On Learning Representations -- ICLR  (2021)

\bibitem{Venkataramanan2021}
Venkataramanan, S., Kijak, E., Amsaleg, L., Avrithis, Y.: Alignmixup: Improving representations by interpolating aligned features. In: IEEE/CVF conference on computer vision and pattern recognition (2022)

\bibitem{Wah2011}
Wah, C., Branson, S., Welinder, P., Perona, P., Belongie, S.: Caltech-ucsd birds 200. Tech. Rep. CNS-TR-2011-001, California Institute of Technology (2011)

\bibitem{Walawalkar2020}
Walawalkar, D., Shen, Z., Liu, Z., Savvides, M.: Attentive cutmix: An enhanced data augmentation approach for deep learning based image classification. In: International Conference on Acoustics, Speech and Signal Processing -- ICASSP (2020)

\bibitem{Xu2023d}
Xu, M., Yoon, S., Fuentes, A., Park, D.S.: A comprehensive survey of image augmentation techniques for deep learning. Pattern Recognition  \textbf{137},  109347 (2023)

\bibitem{Yun2019}
Yun, S., Han, D., Chun, S., Oh, S.J., Yoo, Y., Choe, J.: Cutmix: Regularization strategy to train strong classifiers with localizable features. In: 2019 IEEE/CVF International Conference on Computer Vision -- ICCV (2019)

\bibitem{Zhang2018a}
Zhang, H., Cisse, M., Dauphin, Y.N., Lopez-Paz, D.: mixup: Beyond empirical risk minimization. In: International Conference on Learning Representations (2018)

\bibitem{Zhou2015}
Zhou, B., Khosla, A., Lapedriza, A., Oliva, A., Torralba, A.: Learning deep features for discriminative localization. In: IEEE Conference on Computer Vision and Pattern Recognition -- CVPR (2016)

\end{thebibliography}

\appendix
\section{Training Setup \& Implementation}

\begin{table}
	\caption{Dataset specific parameters}
	\label{tab:dataset-params}
	\centering
	\resizebox{\textwidth}{!}{\begin{tabular}{lrcccc}
			\toprule
			Dataset       & Training Samples & Resolution              & Pretrained & Epochs & Batch Size \\
			\midrule
			CIFAR-100     & 50 000           & $32$ ResNet, $224$ DeiT & \xmark     & $400$  & $100$      \\
			TinyImageNet  & 100 000          & $64$                    & \xmark     & $400$  & $256$      \\
			ImageNet200   & 258 758          & $224$                   & \xmark     & $200$  & $256$      \\
			FGVC-Aircraft & 6 667            & $224$                   & \cmark     & $200$  & $32$       \\
			Stanford Cars & 8 144            & $224$                   & \cmark     & $200$  & $32$       \\
			CUB200        & 5 994            & $224$                   & \cmark     & $200$  & $16$       \\
			\bottomrule
		\end{tabular}}
\end{table}

\begin{table}
	\caption{Data augmentation pipeline}
	\label{tab:data-augment}
	\centering
	\begin{tabular}{ll}
		\toprule
		Training from Scratch & Finetuning           \\
		\midrule

		                      & Resize               \\
		RandomResizedCrop     & RandomResizedCrop    \\
		RandomHorizontalFlip  & RandomHorizontalFlip \\
		RandAugment           & ColorJitter          \\
		\bottomrule
	\end{tabular}
\end{table}

All models are trained using PyTorch.
For datasets trained from scratch (CIFAR-100, TinyImageNet, ImageNet200), we use ResNet with SGD (momentum $0.9$) at a learning rate of $0.1$ and DeiT with AdamW at $10^{-3}$, training for 400 epochs on the smaller datasets and 200 on ImageNet200.
For fine-grained datasets requiring transfer learning (FGVC-Aircraft, Stanford Cars, CUB200), models are initialized from ImageNet-pretrained weights and fine-tuned with reduced learning rates ($0.002$ for ResNet and $10^{-4}$ for DeiT) for 200 epochs.
All runs use a cosine annealing schedule without restarts.
Data augmentation follows two pipelines depending on the training regime: scratch training applies \texttt{RandomResizedCrop}, \texttt{RandomHorizontalFlip}, and \texttt{RandAugment}, while fine-tuning substitutes \texttt{RandAugment} with \texttt{ColorJitter} and prepends a \texttt{Resize} step to resolution $256$.
\Cref{tab:dataset-params,tab:data-augment} present our hyperparameters.

All baselines were implemented by \code{openmixup}\footnote{\url{https://github.com/Westlake-AI/openmixup}}~\cite{Li2022d}.
All training schedule hyperparameters were the same between methods and all method-specific hyperparameters were taken from the original papers and the \code{openmixup} defaults in case a paper did not adequately specify hyperparameters.

\section{Ghost Label Analysis}
\label{sec:ghost-label-analysis}

\begin{figure}[h!]
	\centering
	\resizebox{.8\textwidth}{!}{\includegraphics{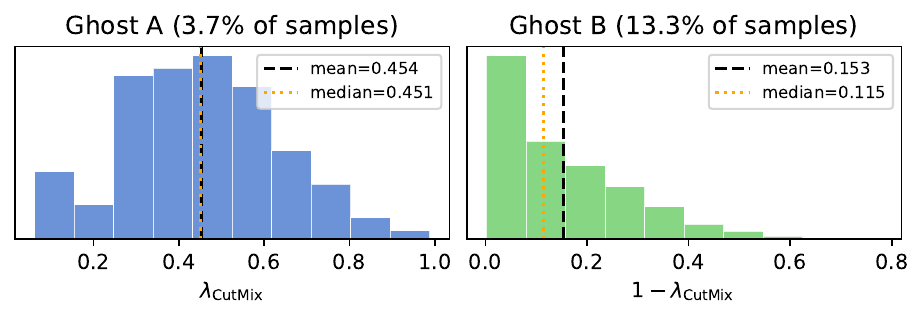}}
	\caption{Distribution of spurious label weights assigned to ghost images.
		Ghost B (pasted patch contributes no visible object pixels) occurs in $13.3\%$
		of samples with a mean spurious weight of $15.3\%$. Ghost A (image A's object
		is entirely covered by the patch) occurs in $3.7\%$ of samples yet still
		receives a mean label weight of $45.4\%$.}
	\label{fig:ghost-label-weights}
\end{figure}

\begin{figure}[h!]
	\centering
	\resizebox{.55\textwidth}{!}{\includegraphics{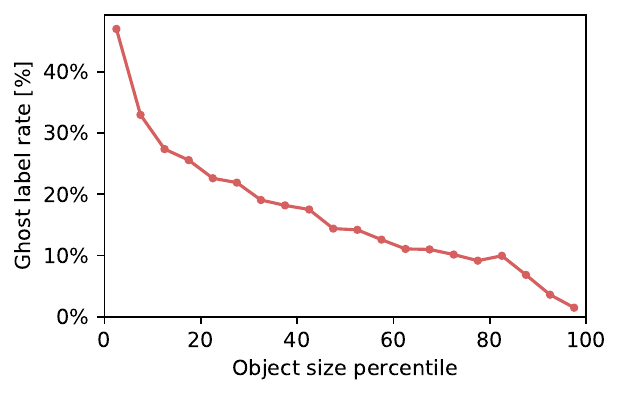}}
	\caption{Ghost label rate as a function of object size percentile.
		For the smallest objects, ghost labels occur in nearly half of CutMix samples,
		decreasing monotonically to zero for large objects that dominate the frame.}
	\label{fig:ghost-vs-objsize}
\end{figure}

We provide a detailed analysis of the ghost label phenomenon.
\Cref{fig:ghost-label-weights} shows the distribution of the spurious label weight assigned in such cases.
Ghost B, where the pasted patch contributes no visible object pixels, occurs in $13.3\%$ of samples, with a mean spurious weight of $15.3\%$.
The complementary Ghost A case, where image A's object is entirely covered by the pasted patch, occurs in $3.7\%$ of samples, yet still receives a mean label weight of $45.4\%$, meaning the model is routinely asked to assign more than half its predictive probability to a class with \emph{no} visual evidence in the image.
\Cref{fig:ghost-vs-objsize} further shows that ghost labels are not uniformly distributed across the dataset:
For the smallest objects (bottom decile), ghost labels occur in approximately a third of CutMix samples, decreasing monotonically to near zero for large objects that fill the frame.
Taken together, these figures show that for small-object classes, CutMix's area-based label assignment is not merely imprecise but often represents completely missing information.
\name{} eliminates ghost labels by construction, assigning a weight of zero whenever no object pixels are visible.

\section{Model Calibration}
\begin{table}
	\caption{Expected Calibration Error (ECE, $\downarrow$) of ResNet models trained from scratch.
		Settings follow Table 1 of the main paper. \name{} achieves the best mean ECE across
		both architectures, improving over CutMix by a large margin.}
	\label{tab:resnet-ece}
	\resizebox{\textwidth}{!}{\begin{tabular}{lcccccccr}
	\toprule
	\multirow{2.5}{*}{Method}                     & \multicolumn{2}{c}{CIFAR-100} & \multicolumn{2}{c}{ImageNet200} & \multicolumn{2}{c}{TinyImageNet} & \multirow{2.5}{*}{Mean} & \multirow{1.5}{*}{latency}                                                    \\
	\cmidrule(lr){2-3}
	\cmidrule(lr){4-5}
	\cmidrule(lr){6-7}
	                                              & ResNet18                      & ResNet50                        & ResNet18                         & ResNet50                & ResNet18                   & ResNet50           &                    & [ms]   \\
	\midrule
	PuzzleMix \cite{Kim2020}                      & $\mathbf{1.10}$               & $4.05$                          & $17.09$                          & $15.86$                 & $9.70$                     & $5.11$             & $8.82$             & 301.00 \\
	Attentive~CutMix \cite{Walawalkar2020}        & $8.53$                        & $12.26$                         & $9.13$                           & $5.86$                  & $7.51$                     & $8.84$             & $8.69$             & 56.36  \\
	SnapMix \cite{Huang2021}                      & $3.10$                        & $5.73$                          & $7.71$                           & $5.79$                  & $8.90$                     & $5.82$             & $6.17$             & 51.49  \\
	\midrule
	Mixup \cite{Zhang2018a}+CutMix \cite{Yun2019} & $11.54$                       & $6.46$                          & $17.79$                          & $17.99$                 & $16.30$                    & $\mathbf{3.44}$    & $12.25$            & 0.67   \\
	Mixup \cite{Zhang2018a}                       & $11.64$                       & $6.03$                          & $15.21$                          & $17.58$                 & $16.62$                    & $5.99$             & $12.18$            & 0.73   \\
	Vanilla                                       & $5.60$                        & $8.20$                          & $\mathbf{4.06}$                  & $4.39$                  & $18.26$                    & $15.46$            & $9.33$             &        \\
	CutMix \cite{Yun2019}                         & $6.14$                        & $11.08$                         & $13.26$                          & $11.84$                 & $6.93$                     & $5.26$             & $9.08$             & 0.61   \\
	FMix \cite{Harris2020}                        & $4.15$                        & $4.12$                          & $10.60$                          & $14.47$                 & $7.10$                     & $5.44$             & $7.65$             & 6.68   \\
	CutMix + decoupled \cite{Liu2023e}            & $2.88$                        & $7.28$                          & $11.68$                          & $10.76$                 & $6.47$                     & $5.48$             & $7.42$             & 0.60   \\
	Alignmixup \cite{Venkataramanan2021}          & $4.00$                        & $\underline{3.61}$              & $\underline{4.33}$               & $\mathbf{1.85}$         & $\mathbf{2.18}$            & $20.81$            & $6.13$             & 1.98   \\
	ResizeMix \cite{Qin2020}                      & $1.93$                        & $5.17$                          & $6.61$                           & $\underline{3.94}$      & $5.40$                     & $7.18$             & $5.04$             & 0.62   \\
	\name/abs + decoupled                         & $\underline{1.74}$            & $5.63$                          & $7.83$                           & $5.19$                  & $3.14$                     & $5.58$             & $4.85$             & 1.01   \\
	\name/rel                                     & $2.56$                        & $3.69$                          & $7.00$                           & $4.97$                  & $3.49$                     & $6.87$             & $4.76$             & 1.02   \\
	\name/rel + decoupled                         & $2.77$                        & $4.53$                          & $5.80$                           & $4.67$                  & $3.18$                     & $7.02$             & $\underline{4.66}$ & 1.00   \\
	\name/abs                                     & $1.79$                        & $\mathbf{3.37}$                 & $7.26$                           & $5.88$                  & $\underline{3.11}$         & $\underline{4.19}$ & $\mathbf{4.27}$    & 1.05   \\
	\bottomrule
\end{tabular}
}
\end{table}

\begin{table}
	\caption{Expected Calibration Error (ECE, $\downarrow$) of DeiT models trained from scratch.
		Settings follow Table 2 of the main paper. \name{} achieves the lowest mean ECE across all
		evaluated methods, including dynamic ones, with \name{} (absolute) reaching a mean ECE
		of $3.27$ compared to $8.75$ for CutMix.}
	\label{tab:deit-ece}
	\resizebox{\textwidth}{!}{\begin{tabular}{lcccccccr}
	\toprule
	\multirow{2.5}{*}{Method}                     & \multicolumn{2}{c}{CIFAR-100} & \multicolumn{2}{c}{ImageNet200} & \multicolumn{2}{c}{TinyImageNet} & \multirow{2.5}{*}{Mean} & \multirow{1.5}{*}{latency}                                                    \\
	\cmidrule(lr){2-3}
	\cmidrule(lr){4-5}
	\cmidrule(lr){6-7}
	                                              & DeiT-Ti                       & DeiT-S                          & DeiT-Ti                          & DeiT-S                  & DeiT-Ti                    & DeiT-S             &                    & [ms]   \\
	\midrule
	SnapMix \cite{Huang2021}                      & $9.07$                        & $7.92$                          & $18.05$                          & $14.44$                 & $4.69$                     & $6.72$             & $10.15$            & 39.98  \\
	Attentive~CutMix \cite{Walawalkar2020}        & $5.96$                        & $5.61$                          & $9.83$                           & $6.84$                  & $5.87$                     & $11.74$            & $7.64$             & 56.10  \\
	TokenMix \cite{Liu2022e}                      & $5.74$                        & $5.18$                          & $13.41$                          & $8.36$                  & $8.01$                     & $\underline{4.45}$ & $7.52$             & 39.91  \\
	PuzzleMix \cite{Kim2020}                      & $9.69$                        & $4.38$                          & $12.17$                          & $2.42$                  & $4.13$                     & $\mathbf{3.81}$    & $6.10$             & 273.28 \\
	\midrule
	Vanilla                                       & $7.56$                        & $19.07$                         & $3.47$                           & $11.19$                 & $17.01$                    & $30.36$            & $14.78$            &        \\
	CutMix \cite{Yun2019}                         & $7.77$                        & $8.92$                          & $13.15$                          & $11.06$                 & $5.61$                     & $5.98$             & $8.75$             & 0.67   \\
	ResizeMix \cite{Qin2020}                      & $7.49$                        & $4.64$                          & $6.55$                           & $5.19$                  & $11.18$                    & $9.57$             & $7.44$             & 1.05   \\
	Mixup \cite{Zhang2018a}+CutMix \cite{Yun2019} & $8.86$                        & $7.33$                          & $11.73$                          & $7.16$                  & $4.08$                     & $5.31$             & $7.41$             & 0.70   \\
	FMix \cite{Harris2020}                        & $3.98$                        & $4.89$                          & $7.38$                           & $5.66$                  & $8.91$                     & $10.63$            & $6.91$             & 6.80   \\
	Mixup \cite{Zhang2018a}                       & $6.49$                        & $6.36$                          & $7.88$                           & $6.02$                  & $4.02$                     & $5.42$             & $6.03$             & 0.73   \\
	CutMix + decoupled \cite{Liu2023e}            & $\underline{3.04}$            & $\mathbf{2.43}$                 & $\underline{2.40}$               & $2.57$                  & $3.88$                     & $13.96$            & $4.71$             & 0.67   \\
	\name/rel + decoupled                         & $4.76$                        & $3.83$                          & $2.80$                           & $1.69$                  & $3.99$                     & $10.53$            & $4.60$             & 1.11   \\
	\name/rel                                     & $3.77$                        & $\underline{3.39}$              & $\mathbf{2.21}$                  & $1.66$                  & $3.60$                     & $9.15$             & $3.96$             & 1.03   \\
	\name/abs + decoupled                         & $\mathbf{3.01}$               & $4.35$                          & $3.26$                           & $\underline{0.77}$      & $\underline{2.84}$         & $7.75$             & $\underline{3.66}$ & 1.01   \\
	\name/abs                                     & $3.18$                        & $3.78$                          & $2.73$                           & $\mathbf{0.68}$         & $\mathbf{2.31}$            & $6.94$             & $\mathbf{3.27}$    & 1.01   \\
	\bottomrule
\end{tabular}
}
\end{table}

Beyond classification accuracy, we evaluate model calibration via the Expected Calibration Error (ECE)~\cite{Guo2017}, which measures the alignment between predicted confidence and empirical accuracy.
Since \name{} produces more semantically accurate soft labels during training, we expect models trained with \name{} to be better calibrated than those trained with standard CutMix.
\Cref{tab:resnet-ece,tab:deit-ece} confirm this hypothesis: \name{} variants achieve the lowest mean ECE for both ResNet and DeiT, outperforming CutMix by a large margin — from $9.08$ to $4.27$ for ResNet and from $8.75$ to $3.27$ for DeiT.
Notably, \name{} also outperforms dynamic methods on calibration despite their more complex mixing procedures.
The improvement is consistent across datasets and architectures, and a slightly larger for absolute mode, hinting at a tradeoff between accuracy and calibration between the two modes.

\end{document}